\documentclass[11pt]{article}
\usepackage{acl2016}
\usepackage{times}
\usepackage{latexsym}
\usepackage{graphicx}
\usepackage{multirow}
\newcommand{\tabincell}[2]{\begin{tabular}{@{}#1@{}}#2\end{tabular}}%
\aclfinalcopy 

\title{Incorporating Loose-Structured Knowledge into LSTM with Recall Gate for Conversation Modeling}

\author{Zhen Xu$^1$, Bingquan Liu$^1$, Baoxun Wang$^2$, Chengjie Sun$^1$, Xiaolong Wang$^{1}$ \\
        $^1$School of Computer Science and Technology, Harbin Institute of Technology, Harbin, China\\
        $^2$Application and Service Group, Microsoft, Beijing, China\\
        $^1${\tt \{zxu, bqliu, cjsun, wangxl\}@insun.hit.edu.cn}\\
        $^2${\tt baoxwang@microsoft.com}
        }

\date{}

\begin{document}

\maketitle

\begin{abstract}
	
	Modeling human conversations
	is the essence for building satisfying chat-bots with multi-turn dialog ability.
	Conversation modeling will notably benefit from domain knowledge since the
	relationships between sentences can be clarified due to semantic hints introduced by knowledge.
	In this paper, a deep neural network is proposed to incorporate background knowledge for conversation modeling.
	Through a specially designed Recall gate,
	domain knowledge can be transformed into the extra global memory of Long Short-Term Memory (LSTM),
	so as to enhance LSTM by cooperating with its local memory to capture the implicit semantic relevance between sentences within conversations.
	In addition, this paper introduces the loose structured domain knowledge base,
	which can be built with slight amount of manual work and easily adopted by the Recall gate.
	Our model is evaluated on the context-oriented response selecting task,
	and experimental results on both two datasets have shown that our approach is promising
	for modeling human conversations and building key components of automatic chatting systems.
	
\end{abstract}

\section{Introduction}
\label{sec:intro}

In recent years, the demand on Chat-bots has changed 
from answering simple questions as a toy to performing smooth open-domain conversations like real humans. 
Some good explorations have been conducted by the open-domain chat-bots like Clever-bot\footnote{http://www.cleverbot.com/ }, etc., 
and their final goal is to reply almost all kinds of queries instead of completing tasks given by users as the normal dialog system\cite{mu2010task} does. 
The ability of communicating like a real human is critical for keeping users' activity, 
thus various additional functions are possible to be introduced during the dialog 
and even the commercial services can be adopted (See Duer\footnote{http://duer.baidu.com/}).

Apparently, the conversation between humans takes \textit{contextual relevance} as the essence with no doubt\cite{ribeiro2015influence}, 
that is, the response should be semantic relevant with both the ``direct'' question and the history contents.
Generally, it is a challenging task for Chat-bots to detect semantic clues of the conversation and provide the context-aware replies. 
For this purpose, the conversation should be well modeled, 
so that the semantic continuation and switching of the context can be sensed 
and the appropriate candidate replies is possible to be further selected.
\begin{figure}[ht]
	\centering
	\includegraphics[width=0.45\textwidth]{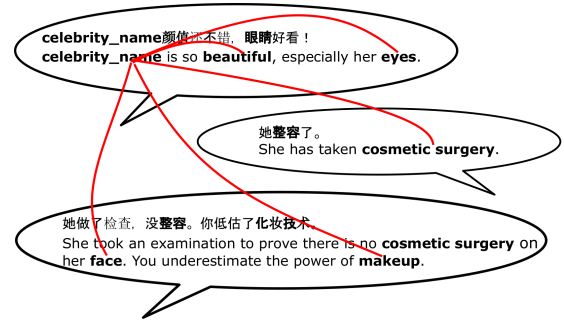}
	\caption{Conversation Example from SNS}
	\label{graph:conversation-example}
\end{figure}

Intuitively, the sequence modeling approaches have great potential to capture the context semantic clues.
Indeed, the Recurrent Neural Network (RNN) and Long Short-term Memory (LSTM)~\cite{hochreiter1997long} based methods have 
already utilized for this task~\cite{sutskever2014sequence,cho2014learning}.
The end-to-end learning ability of such models has already brought promising results 
to some non-trivial problems like sequence-to-sequence mapping~\cite{shang2015neural}.

Obviously, conversation modeling can benefit from human knowledge, only if the proper strategy is taken to adopt knowledge into machine learning models reasonably.
As shown by the example in Figure~\ref{graph:conversation-example},
the semantic clue can be easily captured if the prior knowledge about \textit{celebrity\_name} is involved.
In human conversations, such background knowledge generally takes the role of \textit{global memory}, 
which can be naturally recalled by humans at the right moment. 
Heuristically, it is of great value to simulate the knowledge-supported conversation behavior of human beings, 
by capturing the acting mechanisms of human knowledge in conversation flows, 
so as to recognize semantic relevances in conversations more precisely and further find better responses for creating human-like chat-agents.

This paper proposes a deep neural network to address conversation modeling in the given domain. 
By introducing a new trainable gate to recall the global domain memory,
our deep learning model incorporates background knowledge to enhance the sequence semantic modeling ability of LSTM.
Methodologically, this paper concentrates on learning the implicit working mechanism of global memory in capturing semantic clues in conversations, 
including the functions of recalling and incorporating global memory with short-term memory.
In addition, our model obtains global memory from the loose structured knowledge base.
The knowledge base is built by organizing \textit{entity-attribute} pairs as items 
which can be absorbed into our model by Recall gate directly.
The preparation of such knowledge base requires little amount of manual work, 
by contrast, building complex-structured knowledge (e.g., graph) is indeed not a trivial task.


\section{Related Work}
\label{sec:related}

Previous studies on conversation modeling are mainly task-completion oriented, 
such as act classification, state tracking in spoken field~\cite{jurcicek2011real,williams2013dialog,henderson2013deep}.
With the rapid accumulation of available conversation data from SNS (e.g., Twitter\footnote{https://twitter.com/}, Weibo\footnote{http://weibo.com/}), 
data-driven tasks like response ranking or generation have attracted more attention in this field~\cite{serban2015survey}.

Early studies on conversation modeling focus on \textit{one turn conversation} including one query and a response,
thus heuristically the Information Retrieval (IR) or Statistical Machine Translation (SMT) based methods are taken to get responses.
IR systems are built on movie scripts~\cite{banchs2012iris} or subtitles~\cite{ameixa2014luke} to select responses related to the given questions.
\newcite{ritter2011data} introduce SMT to generate responses by taking query-response pairs as parallel corpus 
and get better performance than the IR approaches.
To improve the readability of responses generated by SMT, 
\newcite{ji2014information} propose an IR framework by introducing learning to rank strategy with the outputs of SMT as matching features.

The multi-turn conversation modeling task becomes even more challenging after taking history contents into consideration,
thus complex approaches with better modeling ability are actually needed.
Since it is natural to consider the conversation as a sequence of short texts,
some studies introduce Neural Network based Sequence-to-Sequence (S2S) framework to address the conversation modeling issue.
\newcite{vinyals2015neural} predict the next sentence based on the given one or several previous sentences with S2S directly.
\newcite{serban2015hierarchical} and \newcite{shang2015neural} take response generation as the decoding process with the 
given distributed representation generated from previous sentences by the encoding process.
The difference between \cite{serban2015hierarchical} and \cite{shang2015neural} lies in that \newcite{shang2015neural}
add a feedback attention mechanism in the encoder-decoder framework. 

\begin{figure*}[ht]
	\centering
	\includegraphics[width=0.9\textwidth]{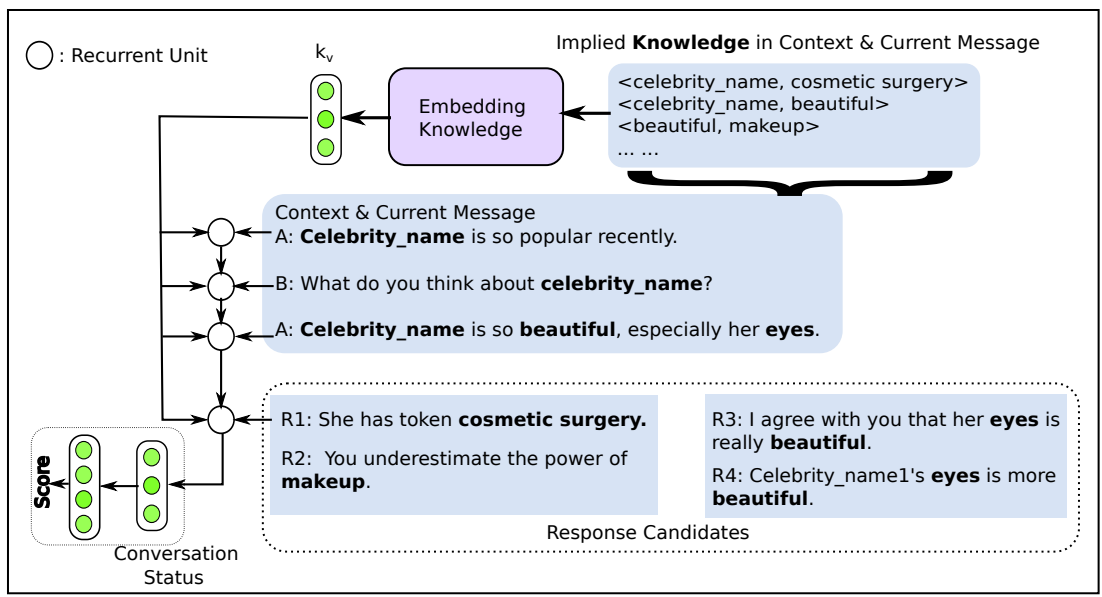}
	\caption{The Framework of Conversation Modeling by LSTM with Recall Gate(r-LSTM).}
	\label{graph:model-arch}
\end{figure*}

Comparing with the generating strategies, 
response selecting approach is an option that avoids low readability naturally.
\newcite{sordoni2015neural} propose an RNN model to evaluate the relevance between contexts, questions and their candidate response, 
and take the relevance as the feature to generate response. 
\newcite{lowe2015ubuntu} and \newcite{kadlec2015improved} also present a ranking strategy on the Ubuntu Dialogue Corpus. 
In their work, an affinity model is used to measure the relevance between the context and a candidate reply, 
and the relevance score is take for response selection.

As pointed by \newcite{vinyals2015neural}, 
the lack of general background knowledge is an obvious limitation of the current conversation modeling approaches.
Noticing that previous studies rarely take background knowledge smoothly in the conversation modeling architectures,
this paper tries to explore the effect of the background knowledge to the Neural Network based models.

\section{LSTM with Recall Gate for Conversation Modeling}
\label{sec:approach}

As mentioned in Section~\ref{sec:intro}, 
our motivation of conversation modeling is to provide reliable evidence to detect users' context-aware queries 
and further select best answers based on the conversation history, 
for building automatic dialog systems.
Apparently, the utterance\footnote{This paper takes utterance as the alias of the sentence in conversations. The usage of this word follows \cite{hurford2007semantics}} 
in a conversation
is very likely to be semantically relevant with both the history \textbf{context} and the \textbf{background knowledge}.
The \textbf{context} is made up of the sequence of utterances appearing in conversation prior to the query and response.
The semantic relevance between utterances implicitly performs as the clue for conversation modeling,
meanwhile, background knowledge offers essential hints to enhance the effect of semantic clues.
In this section, we propose a deep learning framework taking the specially designed Recall gate 
to smoothly integrate knowledge hints into LSTM for capturing such semantic clues.

\subsection{Architecture}
\label{subsec:model-framework}

Figure~\ref{graph:model-arch} illustrates our conversation modeling framework. 
The proposed architecture is composed of sentence modeling, knowledge triggering, and conversation modeling components,
the details of which will be given as follows:

\textbf{Conversation Modeling:} 
Since a conversation can be considered as a sequence of utterances that could be mapped into dense semantic vectors by the sentence modeling component, 
basically we take LSTM based methodology to address the conversation modeling task.
As Figure~\ref{graph:model-arch} shows, 
there are two kinds of inputs: background knowledge and utterance vectors (including context, query and candidate response).
Obviously, for the traditional LSTM, it is difficult to absorb the knowledge vectors properly.

In order to make background knowledge well involved into the deep neural network,
this paper proposes a special Recall gate to enhance LSTM for conversation modeling.
Inspired by our observation of human memory, 
the Recall-Gate is designed to convert the loose-structured domain knowledge to the global memory, 
which cooperates with the local memory in the cell of LSTM to provide evidence to judge whether an utterance is related to the dialog history or not.
The details of LSTM with Recall gate(r-LSTM) will be given in Subsection~\ref{subsec:recall-lstm}.


In our work, 
the target of conversation modeling is to select better responses oriented to the given conversations (composed of the context and a query),
so as to promote user's satisfaction in human-machine conversation. 
For this goal, a \textbf{binary classifier} is set to get the classifying confidence and decide whether a candidate response is related to the current conversation or not.

\begin{figure}[ht]
	\centering
	\includegraphics[width=0.45\textwidth]{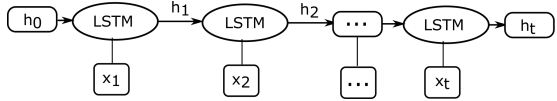}
	\caption{Sentence Modeling Based on LSTM.}
	\label{graph:sent-lstm}
\end{figure}

\textbf{Sentence modeling:} 
As the basis of the entire task, 
sentence modeling aims to provide reasonable semantic representations as inputs of the upper layers.
The RNN based methods, especially LSTM,
have got good performance on sentence modeling recently,
thus our framework adopts LSTM as the component mapping utterances into the real-valued semantic space~\cite{sutskever2014sequence,cho2014learning}.
The process of sentence modeling introduced in this paper is shown in Figure~\ref{graph:sent-lstm}.

The last hidden state $ h_t $
is always taken as a summary of the sentence,
thus we adopt $ h_t $ as the representation of each utterance in conversation.
In our implementation, the LSTM is first pre-trained as a language model for initialization, 
after that, it is tuned according to the end-to-end training of the network in Figure~\ref{graph:model-arch}

\textbf{Knowledge Triggering:}
As shown in the Figure~\ref{graph:model-arch},
contexts will trigger the related background knowledge from the loose-structured knowledge base (KB),
and the related knowledge will be taken as an input signal of our model and transformed into global memory.
We take a simple knowledge trigger method,
assuming that each entity can be correctly located in the KB.

For the given context, 
the process of generating related knowledge vector is as follows:
First, the context is mapped to a bag of entities by matching words to a pre-defined vocabulary described in section~\ref{subsec:Loose Structured Knowledge}.
After that, 
attributes in pairs got from the previous step are ranked by their frequencies.
We select top $N(=10, 20, ... )$ attributes as background knowledge of the given context.
Finally the pairs are mapped to dense vectors~\cite{sukhbaatar2015end} by Equation~\ref{equ:recall-kb}:
\begin{equation}
\label{equ:recall-kb}
kb =  \sum_{i=1}^{i=N}Embedding(attribute_i)
\end{equation}
where $kb$ stands for the knowledge vector to be absorbed by our model.
The embeddings of attributes are pre-trained on a open-domain corpus.

\subsection{LSTM Cell with Recall Gate}
\label{subsec:recall-lstm}

Apparently, prior knowledge is of great value in the process of sequence understanding~\cite{ovchinnikova2012natural}.
Nevertheless, integrating knowledge into deep learning architectures is still a challenging work.
Since knowledge is indeed a global signal,
and it is unwise to simply take knowledge as an additional utterance in conversation understanding,
which is supported by the experimental results in Section~\ref{sec:exp}.
To make domain knowledge perform as a global memory for greater effect,
we design a special model component for LSTM.
\begin{figure}[ht]
	\centering
	\includegraphics[width=0.45\textwidth]{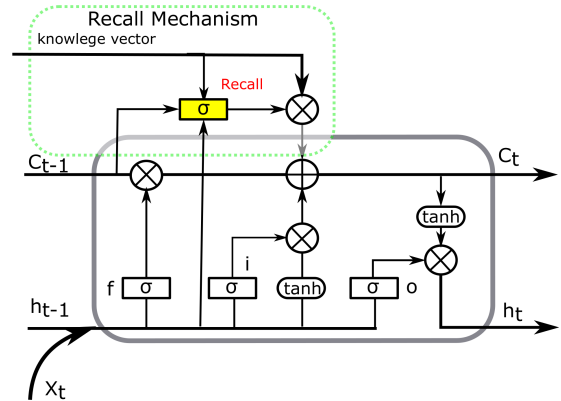}
	\caption{Illustration of the LSTM Cell with Recall Gate.}
	\label{graph:review-lstm}
\end{figure}

Figure~\ref{graph:review-lstm} shows the detail of the LSTM cell with Recall gate (r-LSTM cell).
Basically, our model absorbs the knowledge item embedding converted from the triggering results (as described by Subsection~\ref{subsec:model-framework}), 
and drives it cooperating with the previous sequence state and the present network input to summarize the current status for the next time-step, 
by adding a new recalling mechanism.
Given previous hidden state $ h_{t-1} $, previous memory $ c_{t-1} $ , current input $ x_t $ and background knowledge $ kb $, 
we define the output of Recall gate as follows:
\begin{equation}
\label{equ:recall-1}
\begin{array}{l}
r_t=\\\sigma(W_{ri}[h_{t-1}, x_t] + W_{rc}c_{t-1}
+ W_{rk}kb + b_r)
\end{array}
\end{equation}
where $W_{ri}$, $W_{rc}$ and $W_{rk}$ indicate the connection weights from current input(including previous state and $ x_t $), 
memory cell and background knowledge respectively, 
and $b_r$ is bias of the recall gate.

Generated by Recall gate, $r_t$ indicates the proportion of global knowledge $kb$ taking effect in the determination of current memory cell $c_t$.
This procedure is described by Equation~\ref{equ:recall-2}:
\begin{equation}
\label{equ:recall-2}
c_t =  f_t*c_{t-1} + i_t*c_{input} + r_t*kb
\end{equation}

The current hidden state $h_t$ can be obtained by:
\begin{equation}
\label{equ:recall-3}
h_t = o_t*tanh(c_t)
\end{equation}
where $o_t$ indicates the output gate of LSTM. 
The computations of forget gate, input gate, output gate and input cell are the same as normal LSTM.
From the equations above it can be seen that in our methodology,
external domain knowledge plays a significant role in generating the memory cell of LSTM and influencing the final hidden state consequently. 

In the conversation model illustrated by Figure~\ref{graph:model-arch}, the r-LSTM reads one utterance per time-step and generate a hidden state to represent the current conversation under the condition of global knowledge.
With the multiplication to $r_t$ shown in Equation~\ref{equ:recall-2}, global knowledge $ kb $ including semantic clues takes effect on memory cell.
In this way, the semantic relevance between utterances in conversation process could be sensed well by r-LSTM.
As the higher-level global memory rather than the shallow input, external knowledge will be utilized more effectively in r-LSTM based conversation model.

Comparing with the normal LSTM, the r-LSTM could be more powerful in capturing semantic clues with better effective support of external knowledge because of the recall mechanism of Recall gate.
In other words, the Recall-Gate makes r-LSTM to perform better on incorporating external knowledge.
In contrast to r-LSTM, it is difficult for the general LSTM to involve external knowledge effectively.

\subsection{Loose-Structured Knowledge Base}
\label{subsec:Loose Structured Knowledge}

It is known that 
building a complex-structured knowledge (e.g., WordNet\footnote{http://wordnet.princeton.edu/}, Yago\footnote{http://www.mpi-inf.mpg.de/departments/databases-and-information-systems/research/yago-naga/yago/})
requires large amount of human work.
Obviously, an easily-built knowledge base is quite valuable for applications.
This paper introduces loose-structured knowledge base composed of items with a flexible format ``entity-attribute'', 
in which the \textit{attributes} can be either other entities or the related keywords.

The extraction process of entity-attribute pairs is described as following: 
(1) For a given domain, we extract entity or attributes from the domain-specific corpus by using statistic metrics such as tf-idf, entropy etc.
After that, KL-divergence between domain-specific and general corpora to filter plain words and get entities and attributes for special domain; 
(2) According to the vocabulary composed of entities and attributes, 
counting the frequency of ``entity-attribute'' pair with a slide window.
(3) The final knowledge base is obtained according to the frequency oriented statistics.

\section{Experiment}
\label{sec:exp}

In this part,
our conversation modeling approach will be evaluated on \textit{context-oriented best response selection} task,
which is considered as a binary classification problem as illustrated by Figure~\ref{graph:model-arch},
that is, the goal of our model is to give higher confidence to context related responses.

\subsection{Experimental Setup}
\label{subsec:expsetup}

\textbf{DataSet:}
We execute experiments on two datasets in special domains:
Baidu TieBa Corpus and Ubuntu Corpus\footnote{We have uploaded the Ubuntu dataset to https://www.dropbox.com/s/2fdn26rj6h9bpvl/ubuntu\_data.zip?dl=0 and the Tieba corpus is confidential in the authors' working organization.}.
The Tieba dataset is composed of multi-turn conversations in the \textit{celebrity} domain,
extracted from crawled Tieba threads after several matching and filtering work.
The Ubuntu corpus includes conversations collected from Ubuntu chat rooms focusing on technical supports.

For training the classifier, we adapt sampling method and dataset constructing strategy in~\newcite{lowe2015ubuntu} to generate training and testing set.
For each positive sample in the training set,
one negative sample is prepared.
By contrast, for every positive sample in testing and validation set,
there are 9 negative ones generated correspondingly.
Finally, we construct a training set including 1 million conversations for both Tie and Ubuntu,
and sample 50,000 conversations for validation and testing respectively.
\begin{figure}[ht]
	\centering
	\includegraphics[height=40mm]{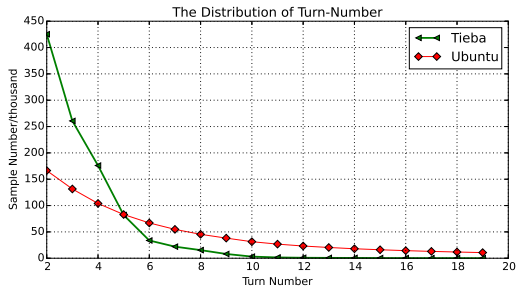}
	\caption{The Distribution of Turn Number.}
	\label{graph:turn-distribution}
\end{figure}

Figure \ref{graph:turn-distribution} shows the distributions of turn numbers of conversations in Tieba and Ubuntu respectively.
Our model needs contexts to trigger background knowledge,
so we select conversations with turn-number ranging from 3 to 7 for experiments.


\textbf{Evaluation Metric:}
Since our model is applied in a classification problem,
and the output confidence scores can be also used for ranking the candidates,
we take \textbf{Accuracy} to measure the performance of classifiers
and introduce \textbf{Recall@k} to evaluate the ranking ability of the approaches.
The metric Recall@k is applied in the work of \newcite{lowe2015ubuntu} for the response selection task.

\textbf{Baseline:}
To illustrate the performance of our model,
we introduce five methods as the baselines,
The descriptions of which are given as follows:

\textit{MLP} and \textit{LSTM}:
Multi-Layer Perceptron(\textit{MLP}) is the first baseline in our experiment.
For our task, this model is designed to take the concatenated sentence vectors of both contexts and candidate responses as input, 
and outputs the binary classification results.
\textit{LSTM} is an intuitive method for our task,
because this model can be directly used to model conversations by taking each utterance embedding as the input of each time-step.

It should be noted that neither of the baselines above considers external knowledge for prediction,
and the architecture of \textit{LSTM} is similar with Figure~\ref{graph:model-arch} but r-LSTM is not involved.

\textit{Affinity Model:}
As assumed by \newcite{lowe2015ubuntu},
context vectors and response vectors generate by RNN or LSTM can be aligned according to a relevance matrix,
which can be formulated as:
\[ p(r|c) = \sigma(c^{T}Mr + b) \]
where $ c $ and $ r $ are the embedding results of context and response given by RNN or LSTM,
and $M$ indicates the relevance matrix.
This methodology achieves the state-of-art results on the Ubuntu corpus without domain knowledge,
thus the reason for introducing this baseline is to show the impact of reasonable knowledge integration.

\textit{MLP+kb} and \textit{LSTM+kb:}
To examine the effectiveness of domain knowledge in conversation modeling,
we incorporate knowledge into the baselines \textit{MLP} and \textit{LSTM} above in a trivial way,
named as \textit{MLP+kb} and \textit{LSTM+kb}.
\textit{MLP+kb} adopts knowledge vector by padding it to conversation vector(concatenated vectors of utterances),
and \textit{LSTM+kb} takes knowledge embeddings as the input of the first time-step.

\textbf{Model Arch and Training Details:}

\textit{Pre-training of word embeddings}: For Ubuntu corpus, we take word vectors from \cite{lowe2015ubuntu} as the initialization of embeddings. 
For Tieba corpus, 
the pre-trained character vectors on threads are utilized. 
All of the approaches (including baselines) in our experiments update word embeddings during the training procedure.
The detailed dimension settings are given in Table \ref{table:dim}.
\begin{table}[h]
	\small
	\begin{center}
		\begin{tabular}{|l|rl|}
			\hline \bf dimension of & \bf Ubuntu & \bf Tieba \\ \hline
			word embedding & 300 & 100 \\
			sentence vector & 200 & 100 \\
			knowledge vector & 200 & 100 \\
			conversation vector & 200 & 100 \\
			\hline
		\end{tabular}
	\end{center}
	\caption{\label{table:dim} Dimension settings of each part in the models.}
\end{table}

For all the models, 
the training process runs until the loss of validation set in current iteration is larger than the previous one.
All the models are trained on a K40m GPU with 12G memory.

\begin{table*}[ht]
	\small
	\centering
	\begin{tabular}{|l|l|c|c|c|c|c|c|}
		\hline
		\multicolumn{2}{|c|}{Method} & Acc & 1 in 2 R@1 & 1 in 10 R@1 & 2 in 10 R@2 & 3 in 10 R@3 & 5 in 10 R@5 \\
		\hline
		\multicolumn{2}{|c|}{MLP}      & 0.6834 & 0.7215 & 0.3118 & 0.4845 & 0.6080 & 0.7848 \\
		\cline{1-8}
		\multicolumn{2}{|c|}{MLP+kb}   & 0.6978 & 0.7660 & 0.3572 & 0.5440 & 0.6670 & 0.8240 \\
		\cline{1-8}
		\multicolumn{2}{|c|}{LSTM}     & 0.6890 & 0.7713 & 0.3696 & 0.5600 & 0.6810 & 0.8316 \\
		\cline{1-8}
		\multicolumn{2}{|c|}{LSTM+kb}  & 0.7140 & 0.7993 & 0.4121 & 0.6041 & 0.7172 & 0.8565 \\
		\cline{1-8}
		\multirow{2}{*}{Affinity Model}
		& \multirow{1}{*}{RNN}  & 0.6826 & 0.7286 & 0.3210 & 0.4896 & 0.6109 & 0.7810 \\
		\cline{2-8}
		& \multirow{1}{*}{LSTM} & 0.6925 & 0.7825 & 0.3814 & 0.5685 & 0.6804 & 0.8353 \\
		\cline{1-8}
		\multicolumn{2}{|c|}{r-LSTM}   & \bf 0.7260 & \bf 0.8172 & \bf 0.4348 & \bf 0.6361 & \bf 0.7527 & \bf 0.8815 \\
		\hline
	\end{tabular}
	\caption{\label{table:tieba-rel} Experiment Results on Tieba Corpus.}
\end{table*}

\begin{table*}[ht]
	\small
	\centering
	\begin{tabular}{|l|l|c|c|c|c|c|c|}
		\hline
		\multicolumn{2}{|c|}{Method} & Acc & 1 in 2 R@1 & 1 in 10 R@1 & 2 in 10 R@2 & 3 in 10 R@3 & 5 in 10 R@5 \\
		\hline
		\multicolumn{2}{|c|}{MLP}      & 0.6837 & 0.7548 & 0.3858 & 0.5343 & 0.6563 & 0.8060 \\
		\cline{1-8}
		\multicolumn{2}{|c|}{MLP+kb}   & 0.7123 & 0.8126 & 0.4840 & 0.6248 & 0.7361 & 0.8645 \\
		\cline{1-8}
		\multicolumn{2}{|c|}{LSTM}     & 0.7162 & 0.8332 & 0.5418 & 0.6957 & 0.7437 & 0.8863 \\
		\cline{1-8}
		\multicolumn{2}{|c|}{LSTM+kb}  & 0.7322 & 0.8679 & 0.6043 & 0.7535 & 0.8087 & 0.9119 \\
		\cline{1-8}
		\multirow{2}{*}{Affinity Model}
		& \multirow{1}{*}{RNN}  & 0.6850 & 0.7618 & 0.3963 & 0.5430 & 0.6732 & 0.8071 \\
		\cline{2-8}
		& \multirow{1}{*}{LSTM} & 0.7238 & 0.8542 & 0.5846 & 0.7214 & 0.7864 & 0.9020 \\
		\cline{1-8}
		\multicolumn{2}{|c|}{r-LSTM}   & \bf 0.7518 & \bf 0.8892 & \bf 0.6493 & \bf 0.7853 & \bf 0.8578 & \bf 0.9324 \\
		\hline
	\end{tabular}
	\caption{\label{table:ubuntu-rel} Experiment Results on Ubuntu Corpus.}
\end{table*}

\subsection{Results and Analysis}
\label{subsec:results}

Table \ref{table:tieba-rel} and Table \ref{table:ubuntu-rel} 
list the experimental results of all the approaches for the context-oriented response selection task,
on Chinese Tieba and English Ubuntu corpus respectively.

From Table \ref{table:tieba-rel} and Table \ref{table:ubuntu-rel}, 
it can be observed that our model (r-LSTM) notably outperforms the baseline methods on both datasets by all the metrics as expected.
Basically, we ascribe the promotions to the fact that our model effectively captures semantic clues with the sequence modeling architecture of r-LSTM,
meanwhile, benefited from smoothly incorporating the loose structured knowledge of the Recall Gate (see Figure~\ref{graph:review-lstm}),
r-LSTM is able to detect implicit semantic hints in both conversation histories and candidate responses.
In addition, the results also shows that 
vectors generated by LSTM based sentence modeling can represent sentences well as described in subsection~\ref{subsec:model-framework}.

The results of Affinity Model shown in Table \ref{table:ubuntu-rel} is lower than those reported in \newcite{lowe2015ubuntu}.
There are to possible explanations for this observation:
First, our experimental dataset extracted from the raw corpus are different from that used in \cite{lowe2015ubuntu},
even though the extraction method is the same as \newcite{lowe2015ubuntu}.
Thus the change of data distribution may influence the performance of this model;
Second, the average length of contexts (by word) in the dataset used in this paper is 57.81, 
by contrast, in \cite{lowe2015ubuntu} the average context length is 108.93.
As mentioned in section \ref{sec:intro},
semantic clues in context are essential for response selection,
so more context could offer more semantic information.
We will open the dataset to other researchers for further comparison.

Ignoring RNN Affinity model, 
the baseline methods shown in the tables can be classified into two categories:
MLP based methods and LSTM based ones.
It's clear that LSTM based methods perform better than the MLP based approaches,
which indicates the ability of capturing semantic clues plays significant role in our task, 
and obviously sequence models have congenital advantage.
Heuristically, explicit or implicit semantic relationships between utterances exist in human-to-human conversations
to keep the continuity of conversations.
Due to the ``memory cell'',
LSTM can model the long-dependent semantic relationships effectively,
so approaches based on LSTM make better understanding of conversations.

\begin{table*}[ht]
	\small
	\begin{center}
		\begin{tabular}{|c|c|c|c|}
			\hline
			\multicolumn{1}{|c|}{context} & \multicolumn{3}{|l|} { \tabincell{l}{I like \textbf{celebrity-name} madly! \\I like him as well! His performance on Spring Festival Gala is very real! \\His model on Spring Festival is very comical. But I still like him.} } \\
			\hline
			LSTM+kb & r-LSTM & Label & Response \\
			\hline
			0.7959	& \bf 0.8676 &	1	 & Me too. This \textbf{uncle} is really \textbf{handsome}! \\
			\hline
			\bf 0.8655	& 0.7555 &	0	 & I always feel that he is old-fashioned. \\
			\hline
			0.3531  	& 0.1172 &	0	 & So embarrassed tonight. \\
			\hline
			\hline \multicolumn{1}{|c|} {context} & \multicolumn{3}{|l|}{\tabincell{l} {Is \textbf{celebrity-name} still a super-star? \\Her \textbf{acting} is poor, isn't it? \\celebrity-name remains at TV acting level, her cinematic feeling is so weak. }} \\
			\hline
			0.6772  & \bf 0.9078 &	1	 & \tabincell{c} {Comparing with \textbf{actresses} in the same period,\\ her \textbf{acting} and \textbf{appearance} are both good.} \\
			\hline
			\bf 0.9278	& 0.8217 &	0	 & I like celebrity-name-1, the real \textbf{actress}. \\
			\hline
			0.5598	& 0.0419 &	0	 & I do not think so, because they just took it out for a walk. \\
			\hline
		\end{tabular}
	\end{center}
	\caption{\label{cases} Samples of best response selection. Entities are anonymous during translation.}
\end{table*}

As shown in Table \ref{table:tieba-rel} and \ref{table:ubuntu-rel},
methods taking account of background knowledge get 3\%-7\% improvements
comparing with the ones without any external knowledge on the best response selection task.
This phenomenon indicates that knowledge is one of the primary factors for conversation understanding.
In detail, there are two aspects influencing the improvement of approaches incorporating domain knowledge:
(1) The quality and quantity of knowledge.
Benefited from the characteristics of loose structured knowledge,
we can build a knowledge base with large amount of domain specified information and update it with little manual work.
Even the quality of such loose structured knowledge is lower than the knowledge base built with much human efforts (like Knowledge Graph),
our model can still effectively utilize it.
(2) The fusion strategy between knowledge and utterances in conversation modeling.
Our model r-LSTM obtains up to 2-4\% improvement by all the metrics than ``LSTM+kb'' on both datasets.
This demonstrates the importance of strategies for incorporating external knowledge in conversation modeling.
Moreover, there are semantic gaps between loose structured knowledge and sentences,
it's troublesome for methods like ``LSTM+kb'', 
to overcome these gaps for methods taking knowledge as an additional utterance directly.

The \textit{Recall@3} of our model on Tieba and Ubuntu dataset is 75.27\% and 85.78\% respectively,
that is, most best responses can be recalled in the top three candidate responses (ranked by confidence).
This can be attributed to the recall mechanism of r-LSTM involving background knowledge as global memory.
As the global input, external domain knowledge can be recalled by r-LSTM
and influence the local memory at right moments in the whole conversation process.
The introduction of background knowledge provide essential semantic hints 
thus enhances the ability to detect the semantic relevances between sentences.

Due to the limitation of quality and average length of contexts in Tieba,
the models' overall performances are a little lower than those on the Ubuntu,
as shown in Table \ref{table:tieba-rel} and \ref{table:ubuntu-rel}.
According the distribution of turn numbers shown in Figure \ref{graph:turn-distribution},
there are few conversations with more than 5 turns in the Tieba corpus.
Consequently, there are less history contents involving semantic clues for selecting response as mentioned in section \ref{sec:intro}.

\subsection{Case Study}
\label{subsec:case-study}

In order to further show the working mechanism in our model intuitively,
we give two cases comparing our r-LSTM and LSTM+kb in Table~\ref{cases}.
For better understanding, 
we translate both the contexts and the candidate responses into English from Chinese.
As shown by Table \ref{cases},
each case contains the context composed of three utterances and three candidate responses with labels and predicted scores.
The label 1 indicates the true response to the given context.
Scores represent the confidence of responses as the best one.

From the table it can be seen that r-LSTM gives highest score to best response,
while ``LSTM+kb'' offers unsatisfied results.
Both methods shown in \ref{cases} involved background knowledge, 
so this result is caused by the effectiveness of utilizing knowledge.
In our r-LSTM architecture, 
background knowledge is considered as global memory and can be recalled in the conversation modeling process,
and our Recall Gate could also build semantic relationships of utterances validly.
Taking knowledge as input directly, 
LSTM+kb makes less use of knowledge in the modeling process because of the semantic gap between sentences and knowledge items.
Furthermore, based on the observation of human conversations, 
it is reasonable to consider background knowledge as global signal of the neural network,
instead of an additional input as LSTM+kb does.
The large range of confidence scores given by r-LSTM also shows
that our model does well on both selecting best responses and recognizing inappropriate responses.

In both the two samples,
the second response also gets high confidence score,
and LSTM+kb even takes it as the best one.
Actually, such responses are general responses without any key points.
Even though they can be taken to reply the queries,
such general answers are not promising because less dialog turns are expected after them.
Thus distinguishing general and best responses is of great value for chat-agents,
and our model have potential in this scenario.

\section{Conclusions and Future Work}
\label{sec:con}

In this paper,
we proposed a deep learning architecture to incorporate loose structured knowledge for end-to-end conversation modeling. 
Our approach shows good potential on the context-oriented response selecting task.

The contributions of this paper can be summarized as follows:
(1) By investigating the influence of domain knowledge on conversation understanding,
we present a LSTM framework with a designed Recall gate to utilize knowledge smoothly and effectively.
Transforming knowledge into global memory, 
the Recall gate enables LSTM to integrate global memory into sequential local memory to conduct enhanced conversation modeling.
(2) To guarantee the flexibility of domain knowledge base for practical usage,  
this paper introduces the loose-structured knowledge base organized as ``entity-attribute'' pairs, 
which can be extracted easily from raw corpora.
The knowledge can be directly absorbed by the Recall gate after being triggered according to dialog contexts.

Our future study will be conducted along the following directions: 
first, we will continue exploring the working mechanism of global memory to refine our Recall gate;
Second, the utilization of our model for open-domain conversation modeling will be investigated.

\bibliography{acl2016}
\bibliographystyle{acl2016}

\end{document}